\documentclass[runningheads]{llncs}
\usepackage{graphicx}
\usepackage{comment}
\usepackage{amsmath,amssymb} % define this before the line numbering.
\usepackage{color}

\begin{document}
% \renewcommand\thelinenumber{\color[rgb]{0.2,0.5,0.8}\normalfont\sffamily\scriptsize\arabic{linenumber}\color[rgb]{0,0,0}}
% \renewcommand\makeLineNumber {\hss\thelinenumber\ \hspace{6mm} \rlap{\hskip\textwidth\ \hspace{6.5mm}\thelinenumber}}
% \linenumbers
\pagestyle{headings}
\mainmatter
\def\ECCVSubNumber{0000}  % Insert your submission number here

\title{Photo-Realistic Video Prediction on Natural Videos of Largely Changing Frames}

% INITIAL SUBMISSION 
\begin{comment}
\titlerunning{ECCV-20 submission ID \ECCVSubNumber} 
\authorrunning{ECCV-20 submission ID \ECCVSubNumber} 
\author{Anonymous ECCV submission}
\institute{Paper ID \ECCVSubNumber}
\end{comment}
%******************

% CAMERA READY SUBMISSION
%\begin{comment}
\titlerunning{Photo-Realistic Video Prediction}
% If the paper title is too long for the running head, you can set
% an abbreviated paper title here
%
\author{Osamu Shouno\inst{1}\orcidID{0000-0001-7385-3280} }
\authorrunning{O.Shouno}
% First names are abbreviated in the running head.
% If there are more than two authors, 'et al.' is used.
%
\institute{Honda Research Institute Japan Co., Ltd., Wako, Saitama, Japan} 
%\email{osamu.shouno@gmail.com}\\
%\email{shouno@jp.honda-ri.com}\\
%\url{https://www.jp.honda-ri.com/en/} 
%\end{comment}
%******************
\maketitle

\begin{abstract}
Recent advances in deep learning have significantly improved performance of video prediction. 
However, state-of-the-art methods still suffer from blurriness and distortions in their future predictions, especially when there are large motions between frames.
To address these issues, we propose a deep residual network with the hierarchical architecture
where each layer makes a prediction of future state at different spatial resolution, 
and these predictions of different layers are merged via top-down connections to generate future frames. 
We trained our model with adversarial and perceptual loss functions, and evaluated it on a natural video dataset captured by car-mounted cameras. 
Our model quantitatively outperforms state-of-the-art baselines in future frame prediction on video sequences of both largely and slightly changing frames. 
Furthermore, our model generates future frames with finer details and textures that are perceptually more realistic than the baselines, especially under fast camera motions.  %(149 words)

\end{abstract}

\section{Introduction}
The capability of an artificial agent to forecast how a visual environment can evolve in the future can be applied to robotics, autonomous driving, health-care, and action recognition.
Video prediction is the task of predicting future frames given past video frames. 
A model is trained from unlabeled videos and learns representations for object and scene structures and transformations of their appearances.
However, unsupervised learning of visual prediction on natural videos is challenging because of the diversity of objects and backgrounds in a scene, 
various sizes of object appearance, occlusions, disocclusions, camera movements, and other dynamic scene changes between frames.

Deep learning technologies have enabled a video prediction model to learn from large-scale, unlabeled real-world videos and improved its performance of future prediction \cite{ranzato2014video,srivastava2015unsupervised,finn2016unsupervised,Mathieu16,vondrick2017generating,Lotter17,liu2017video,villegas2017learning,Byeon_2018_ECCV,kwon2019predicting,Gao_2019_ICCV}. 
One of the major unsolved problems in video prediction is blurry predictions. 
Two factors have been addressed as the primary causes of this artifact. 
Firstly, simple pixel-wise loss functions based on the mean squared error (MSE) or mean absolute error (MAE), 
which are often used for objective functions of video prediction models, 
are not capable of capturing long-range correlations among pixels which are characteristic to natural images \cite{theis2012mixtures,oord2016pixel,kalchbrenner2017video}, 
nor accounting for uncertain futures \cite{Mathieu16,vondrick2016generating,walker2016uncertain}.
Other objective functions, such as adversarial loss and perceptual loss functions, are shown to be effective at reducing blurry artifacts 
in the predicted frames \cite{Mathieu16,vondrick2017generating,kwon2019predicting,Gao_2019_ICCV}, 
but not as effective as it is in other image generation tasks such as the super-resolution \cite{johnson2016perceptual,Ledig_2017_CVPR}.
Secondly, network architectures can be inappropriate for video prediction. 
Byeon et al. \cite{Byeon_2018_ECCV} discussed the blind spot problem 
in which a video prediction model is not capable of accessing to entire available past information for predicting each pixel, increasing model uncertainty. 
Empirically, removing the blind spots improved performance of video prediction \cite{Byeon_2018_ECCV} .
\begin{figure}
\centering
\includegraphics[height=10.5cm]{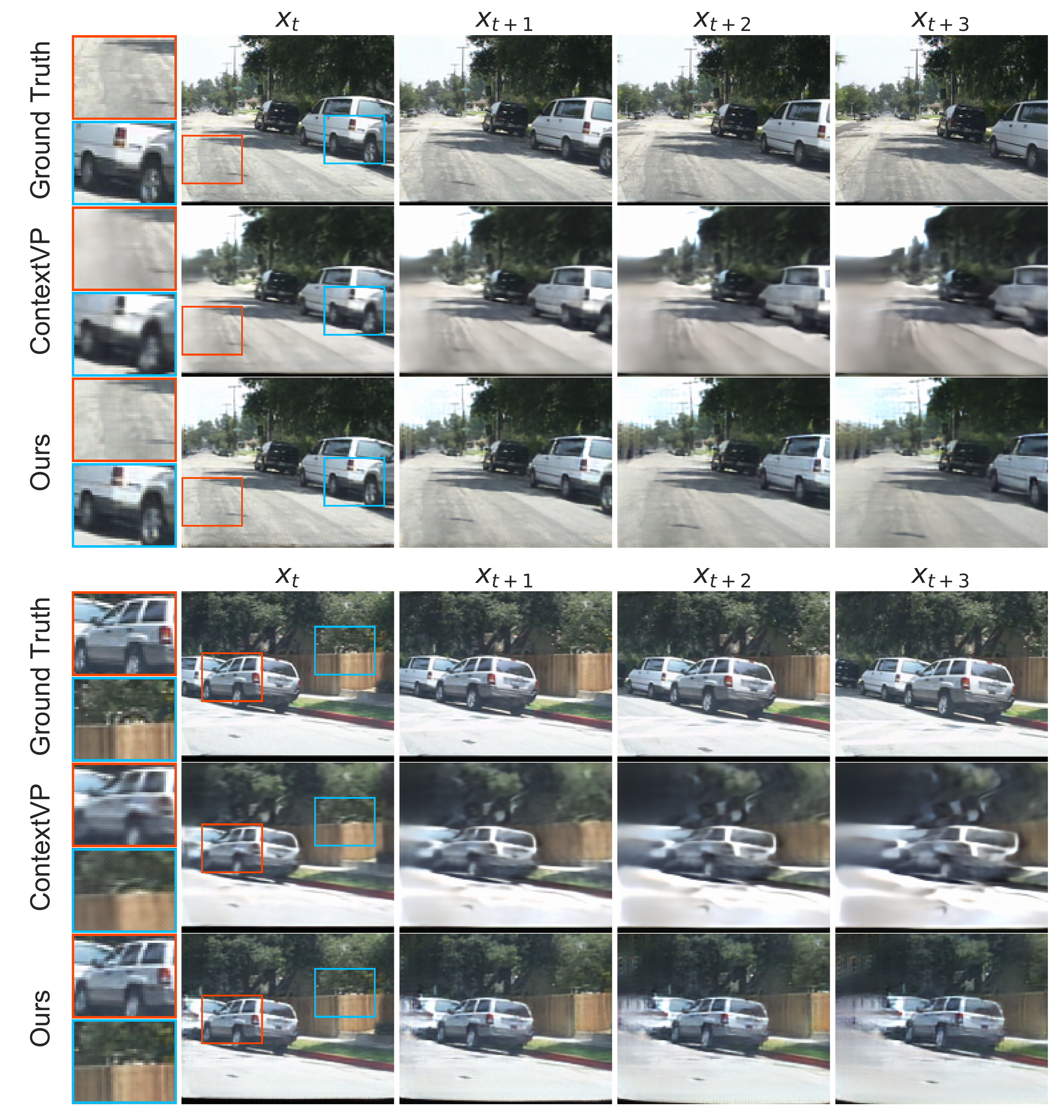}
\caption{Qualitative comparison of predicted frames between the state-of-the-art model, ContextVP \cite{Byeon_2018_ECCV}, and our model (GAN-VGG). 
ContextVP and our models are trained for next-frame prediction on the KITTI dataset, 
and tested to recursively predict future frames on the Caltech Pedestrian dataset.
As an input for the prediction, ContextVP and our model take 10 and 8 frames, respectively.
Our model reproduces finer details and textures compared to ContextVP as denoted by bounding boxes.
Results are best viewed in color with zoom}
\label{fig:multi-frame-examples}
\end{figure}

However, current state-of-the-art video prediction models still suffer from blurriness and distorted object shapes in their predictions, 
especially when there are large camera movements \cite{Byeon_2018_ECCV,Gao_2019_ICCV} as illustrated in Fig. \ref{fig:multi-frame-examples}.  
Additionally, their next-frame predictions often lack fine details and textures and are overly smooth.
Even if such prediction errors seem to be slight, they can quickly accumulate when recursively predicting further future frames due to the discrepancy between training and prediction, 
leading to large distortions of objects. 

In this work, we propose a framework for video prediction 
that takes advantage of the adversarial and perceptual loss functions to generate accurate and photo-realistic future frames. 
In order to fully utilize the potential of these two loss functions, a neural network model for video prediction should be sufficiently powerful and expressive 
to be capable of modeling a variety of structures of objects,  
and  various types of spatio-temporal transformations of their appearances not only at a low-frequency level but also at a high-frequency level. 
Inspired by the architecture of the PredNet \cite{Lotter17}, 
we propose a deep residual network with hierarchical architecture for video prediction 
where each layer models object and scene structures and their spatio-temporal transformations for making a prediction of their future state at different spatial resolution.
These predictions of different layers are sequentially merged via top-down connections to generate future frames. 
Our model is trained end-to-end with the adversarial loss function computed using the generative adversarial network framework  (GAN) \cite{goodfellow2014generative,radford2015unsupervised} 
and perceptual loss functions based on differences between learned visual feature representations of pre-trained deep CNNs \cite{gatys2015texture,BrunaSL15,johnson2016perceptual,Ledig_2017_CVPR}.
The main contribution of this paper is the development of a new method for predicting accurate and perceptually realistic future frames of natural videos even if there are large motions between frames.  
Examples of perceptually realistic future frames are shown in Fig. \ref{fig:multi-frame-examples}. 

\section{Related Work}
Recent work on unsupervised learning of video prediction has tackled next-frame and longer-term prediction in synthetic and real-world videos.
Srivastava et al. \cite{srivastava2015unsupervised} proposed a recurrent neural network model featuring LSTM for predicting future frames.
Oh et al. \cite{oh2015action} proposed encoder-decoder neural network models for predicting future frames conditioned on actions.
These models suffered from blurry predictions due to the pixel-wise MSE loss functions that inherently generate blurry results.

Later models have challenged to solve blurry predictions with different approaches.
One approach focuses on loss functions, especially on adversarial loss functions computed from a GAN framework \cite{goodfellow2014generative}.
Mathieu et al. \cite{Mathieu16} proposed a GAN-based framework with multi-scale convolutional network for predicting sharper future frames.
Vondrick et al. \cite{vondrick2017generating} defined a GAN-based model featuring differential transformers for generating future frames.
Liang et al. \cite{Liang_2017_ICCV} showed that shared internal representations for future-frame prediction and optical flow estimation improved future frame prediction.
Kwon and Park \cite{kwon2019predicting} proposed a GAN-based solution featuring retrospective cyclic constraints.
Another approach considers effective usage of optical flows.
Liu et al. \cite{liu2017video} proposed a model that explicitly represents optical flow of pixels for generating future frames with warping function. 
Their results are sharp, however, tend to errors where flow predictions are incorrect or ill-posed.
Gao et al. \cite{Gao_2019_ICCV} augmented the flow-based solution with the inpainting module that hallucinates pixels for erroneous regions and with the perceptual loss functions.
Other approaches focus on network architectures.
Lotter et al. \cite{Lotter17} proposed the PredNet architecture for unsupervised learning of videos based on the predictive coding, and demonstrated on challenging natural videos recorded from car-mounted cameras.
Byeon et al. \cite{Byeon_2018_ECCV} pointed out the blind spot problem in network architecture that prevents effective usage of given past information for predicting each pixel.
These approaches have improved performance of video prediction, however, still suffer from blurry, inaccurate predictions that lack fine details and textures.

Our approach considers both loss functions and network architecture for photo-realistic video prediction. 
Prior works on the single image super resolution \cite{johnson2016perceptual,Ledig_2017_CVPR} show that 
the adversarial and perceptual loss functions are effective for generating photo-realistic images with fine details and textures.
They adopted a deep residual neural network with an encoder-decoder architecture in which most of residual blocks are used for processing at a lower spatial resolution. 
However, video prediction needs to handle a more diverse range of image transformations than the single image super-resolution.
The works mentioned above \cite{kwon2019predicting,Gao_2019_ICCV} used frameworks very similar to the super-resolution \cite{johnson2016perceptual,Ledig_2017_CVPR}, but their results are not qualitatively comparable to those of the super-resolution.
We therefore focus on enhancing network capacity of a video prediction model 
by introducing a hierarchical architecture for parallel multi-scale spatio-temporal processing
to balance between network capacity and constraints imposed by these loss functions.

Our network architecture is a solution for the blind spot problem similar to the multi-scale convolutional network \cite{Mathieu16}, but different in three points. Firstly, instead of conventional image precessing methods \cite{Mathieu16}, our model uses residual neural networks for down- and up-sampling which can be optimized through the back-propagation. 
Secondly, we adopt a single adversarial loss for a finally generated frame instead of scale-wise adversarial losses.
Finally, we use three-dimensional convolutions for spatio-temporal processing instead of two-dimensional ones.
Another strong solution for the blind spot problem is exhaustive scanning of a video frame sequence along not only temporal but also all spatial directions by recursive application of convolutional LSTMs \cite{Byeon_2018_ECCV}. This solution can be optimized through the back-propagation-through-time which consumes a large amount of memory proportional to the length of sequence, hampering increasing the model complexity due to the limited amount of computational resources of commercially available devices.

\section{Proposed Method}
Our goal is to learn a video prediction model which can predict future frames that are sharp and perceptually realistic given a sequence of past frames from natural videos,  
even when there are large scene changes between frames.

\newcommand{\argmin}{\rm arg \it \mathop{\rm min}\limits}
\subsection{Objective function} 
Our model adopts a framework of a generative adversarial network (GAN) that consists of a generator network $G$, and discriminator network $D$. 
Here, $G$ is a video prediction model that learns to map a given input video sequence of $t$ frames $\mathbf{x_{0:t}}=\{x_0, \dots x_{t-1}\}$ to the predicted future frames  $\mathbf{\hat{x}_{t:T}}$, 
while $D$ is trained to distinguish the predicted frames $\mathbf{\hat{x}_{t:T}}$ from real videos $\mathbf{x_{t:T}}$.
The GAN optimization with the hinge version of the standard adversarial loss functions \cite{miyato2018spectral} involves the following objectives:
\begin{align}
G^* &= \argmin_{G}   \lambda_1 \mathcal{L}_{adv}^G + \lambda_2 \mathcal{L}_{MAE} + \lambda_3 \mathcal{L}_{VGG}  \\
D^* &= \argmin_{D} \mathcal{L}_{adv}^D 
\end{align}
The objective function for the generator is a weighted combination of an adversarial loss function and two types of reconstruction loss functions.
The hinge-version adversarial loss functions are defined as:
\begin{align}
\mathcal{L}_{adv}^G &= -  \mathbb{E}_{\mathbf{x_{0:t}}, \mathbf{z}} \left [ D(G(\mathbf{x_{0:t}, z}),\mathbf{x_{0:t}}) \right ] \\
\begin{split}
\mathcal{L}_{adv}^D &= \mathbb{E}_{\mathbf{x_{0:t}}, \mathbf{x_{t:T}}}\left [\rm max  (0, 1 -\it D(\mathbf{x_{t:T}, x_{0:t}}) ) \right ] \\
& \quad +   \mathbb{E}_{\mathbf{x_{0:t}}, \mathbf{z}} \left [ \rm max \left(0,  1+\it D(G(\mathbf{x_{0:t}, z}),\mathbf{x_{0:t}}) \right ) \right ]
\end{split}
\end{align}
where $\mathbf{z}$ is a noise vector.

The reconstruction loss functions include the pixel-wise, mean absolute error loss (\bf{MAE loss}\rm)  function defined as follows:
\begin{align}
\mathcal{L}_{MAE} = \| \mathbf{\hat{x}_{t:T}} - \mathbf{x_{t:T}} \|_1^1
\end{align} 
Along with the MSE, this is one of the most widely used loss functions for future video prediction on which the state-of-the-art methods rely \cite{Byeon_2018_ECCV,Lotter17}. 
Although solutions of MAE/MSE optimization problems achieved particularly high scores in the traditional metrics such as the peak signal-to-noise ratio (PSNR) and Structural Similarity Index Measure (SSIM),
they often result in blurry images which lack high-frequency details and thus \it perceptually \rm distant from the target frames for human observers.

Recent works show that instead of a pixel-wise loss function, 
a perceptual loss function based on differences between learned visual feature representations of pre-trained deep CNNs  
is very effective for generating high-quality images with fine details \cite{gatys2015texture,BrunaSL15,johnson2016perceptual,Ledig_2017_CVPR}.
We therefore adopt the second reconstruction function, a perceptual loss function based on cosine distances between visual feature representations of pre-trained CNNs which is found to correlate well with human perception \cite{zhang2018unreasonable}.
Let $\phi^l$ be the visual feature representations extracted from the last convolutional layer of the $l$-th block of VGG-16 \cite{SimonyanZ14a}.
We define the perceptual loss function,  \bf{VGG loss}\rm, as follows: 
\begin{align}
\mathcal{L}_{VGG} = \sum_{l}  \frac{1}{H_l W_l} \sum_{h,w} \left\| \frac{\phi_{h,w}^l(\hat{x})}{\| \phi_{h,w}^l(\hat{x}) \|_2^1} - \frac{\phi_{h,w}^l(x)}{\| \phi_{h,w}^l(x) \|_2^1} \right\|_2^2 
\end{align}
where $H_l$ and $W_l$ denote the dimensions of the $l$-th feature representations.

\begin{figure}
\centering
\includegraphics[height=8.75cm]{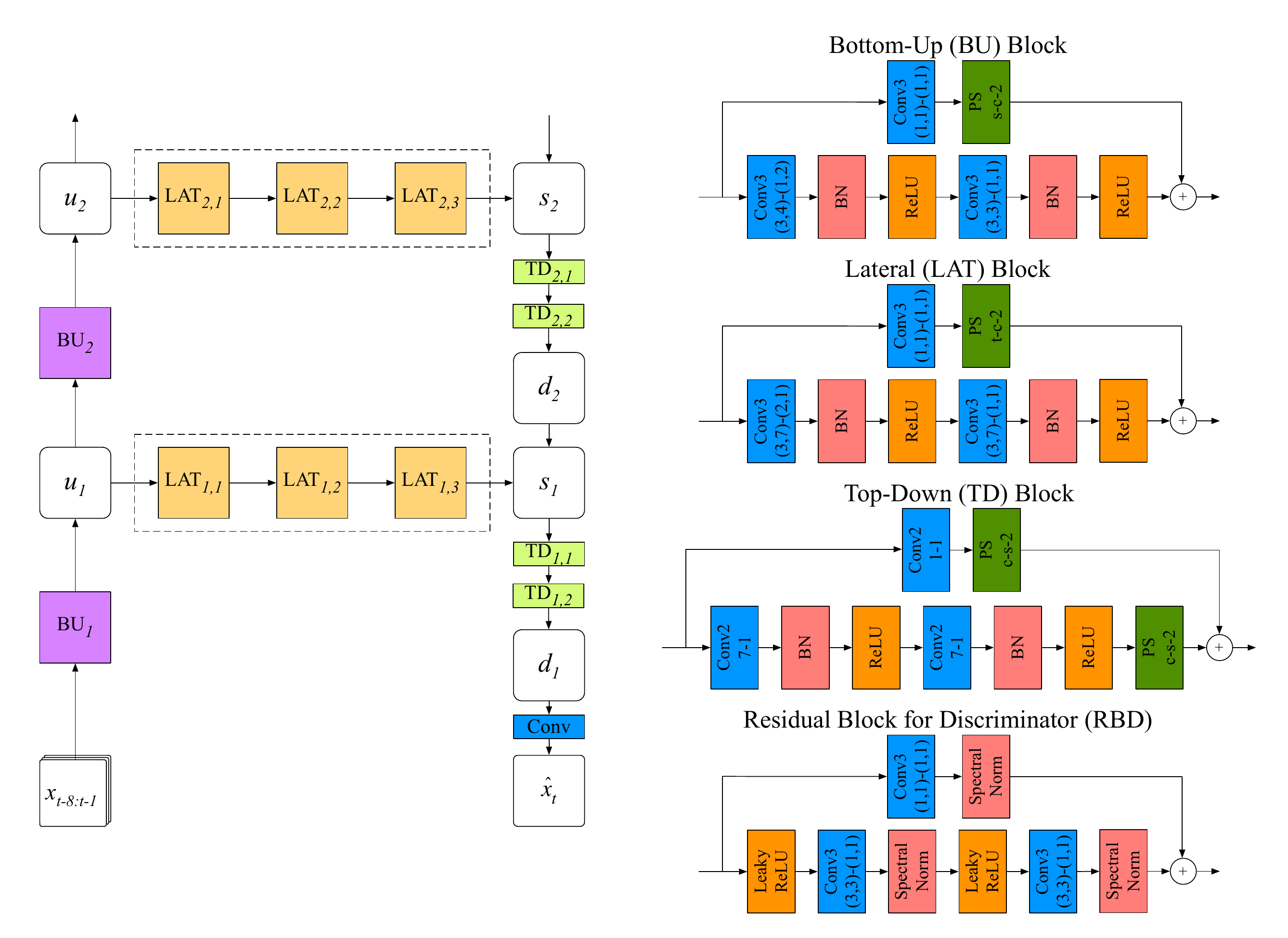}
\caption{Network architecture of the generator network and its building blocks. 
Left: Illustration of information flow within two layers of the generator. Each layer consists of the bottom-up block (BU), a series of the lateral blocks (LAT) and the top-down blocks  (TD).
Right: Designs of the building blocks. 
Conv3 ($k_t$, $k_s$)-($s_t$, $s_s$): $k_t$ and $k_s$ ($s_t$ and $s_s$) indicate kernel sizes (strides) of temporal and spatial dimensions, respectively. 
Conv2-$k$-$s$: $k$ and $s$ indicate kernel size and stride of a spatial convolution.
BN, and Spectral Norm indicate the batch normalization \cite{ioffe2015batch} and spectral normalization \cite{miyato2018spectral}, respectively. 
PS $d0$-$d1$-$r$ indicates the pixel shuffler \cite{Shi_2016_CVPR} 
that upscales an input array along its $d1$ dimension from interspersed $d0$ layer by subpixel array sampling with the upscaling factor $r$. 
Consider that the shape of an input array is $(c, t, h, w)$ where $c, t, h$, and $w$ indicate channel, time, height and width dimensions.  
Similar to $d0$ and $d1$, we use $c$, $t$, or $s$ to indicate channel, time, and space ($h$ and $w$)  dimensions of an input array, respectively}
\label{fig:example}
\end{figure}
\subsection{Network architecture}
The generator network consists of a series of repeating stacked modules which have spatial receptive fields of increasing size
and communicate via bottom-up and top-down connections. 
Briefly, each module of the network consists of three basic parts: a bottom-up block (BU), a sequence of lateral blocks (LAT) and top-down blocks (TD). 
These network blocks are sequences of a 3-dimensional/2-dimensional convolutional layer, batch normalization \cite{ioffe2015batch}, 
ReLU activation function and/or  pixel shuffler \cite{Shi_2016_CVPR}  with a skip connection. 
The bottom-up block of the $l$-th module, $\rm BU\it_{l}$, receives a bottom-up input $u_{l-1}$ from the lower module and computes $u_{l}$,  
a spatially-downscaled representation of the input, and then passes it to the higher module.
\begin{align}
  u_l = \rm BU\it_{l}(u_{l-\rm1})
\end{align}
The bottom-up input to the lowest module, $u_0$, is a sequence of $t$ video frames $\mathbf{x_{0:t}} \in \mathbb{R}^{3 \times t \times h \times w}$. 
We use a fixed value 8 for $t$ in this paper.
$u_{l}$ is also passed forward through a series of lateral blocks ($\rm LAT\it_{l,\rm1}$, $\rm LAT\it_{l,\rm2}$, $\rm LAT\it_{l,\rm3}$) to predict $s_{l}$, 
a next-frame state at the spatial scale of the module, along with top-down input, $d_{l+\rm1}$, from the higher module. 
\begin{align}
  s_l = [\rm LAT\it_{l,\rm3\it}(\rm LAT\it_{l,\rm2\it}(\rm LAT\it_{l,\rm1\it}(u_l))), d_{l+\rm1}]
\end{align}
Then $s_{l}$ is spatially up-sampled through the top-down blocks ($\rm TD\it_{l,\rm1\it}, \rm TD\it_{l,\rm2\it}$) to generate a top-down output to the lower module.
\begin{align}
  d_l = \rm TD\it_{l,\rm2\it}(\rm TD\it_{l,\rm1\it}(s_l))
\end{align}
The top-down output from the bottom module ($l$=1), $d_1$, is further passed through a convolutional layer and sigmoid function to predict the next-frame image. 
We use four modules whose number of channels are (64, 128, 256, 512) for the generator network.
Random noises are injected at the top-down blocks of upper two modules via dropout before applying ReLU activation function as \cite{Isola_2017_CVPR}.

The discriminator network consists of a series of 5 repeating blocks of the residual blocks for the discriminator (RBD) followed by spatial average pooling. 
Number of RBDs and their channels are (1, 2, 2, 2, 2) and (64, 128, 512, 1024, 2048), respectively.
An input to the discriminator network is a predicted next frame $\mathbf{\hat{x}_{t}}$ or its ground truth $\mathbf{x_{t}}$ 
concatenated to the corresponding input frame sequence to the generator $\mathbf{x_{0:t}}\in \mathbb{R}^{3 \times t \times h \times w}$.
The last spatial average pooling is followed by temporal average pooling, 
and the discriminator finally tries to classify if each $N \times M$ %$4 \times 5$ 
patch in a frame is \it real \rm or \it predicted \rm \cite{Isola_2017_CVPR}.

\section{Experiments}
\subsection{Dataset}
We evaluate the proposed approach on car-mounted camera video prediction. 
We use two famous datasets, the KITTI dataset \cite{geiger2013vision} and the Caltech Pedestrian dataset \cite{dollar2009pedestrian}, 
which were taken from vehicles moving around urban areas in Germany and Los Angels, respectively.  
They consist of diverse and complex visual motions of a variety of real-world objects (vehicles, pedestrians, bicycles, roads, buildings, trees, etc.) at different scales 
due to a combination of objects' own movements and camera movements.
Compared to other popular, real-wold datasets, such as Human3.6M and UCF-101 datasets, which have static background,
these datasets are characterized by their broad range of background motions, 
from a static background when an ego-vehicle stops at a red light to large background motions when an ego-vehicle turns right or left at a crossing.

The model is trained on the KITTI dataset and tested on the Caltech Pedestrian test dataset as proposed by \cite{Lotter17}. 
Every ten input frames from 
the training subdivision \cite{Lotter17} of 
the KITTI dataset are sampled for training (about 41K frames as total). 
Note that our models use the last 8 frames as input from the sampled 10-frame-length sequence.
Frames from both datasets are center-cropped and down-sampled to 128 $\times$ 160 pixels. 
In the case of the Caltech Pedestrian dataset whose sampling rate is 3 times higher than the KITTI dataset, 
we beforehand temporally down-sample all sequences of the Caltech Pedestrian dataset to match their sampling rate to that of the KITTI dataset.
All pixel values are normalized to [0,1]. 

\subsection{Training details}
We trained all our networks on a NVIDIA Tesla V100 GPU with 32GB onboard memory with the KITTI dataset.  
Input sequences were horizontally flipped with the probability of 0.5 for data augmentation.
For optimization, we used Adam \cite{KingmaB14} with $\beta_1=0.0$, $\beta_2=0.9$ and a batch size of 8. 
In order to avoid unstable optimization and suboptimal solutions, 
we trained the model end-to-end in three phases.
In the first phase, the discriminator was disabled and the generator was trained to predict the next frame 
given 8 past frames with the MAE and VGG losses with a learning rate $\alpha=10^{-4}$, $\lambda_2=1000$ and $\lambda_3=400$.
Then in the second phase, the discriminator was trained with a learning rate  $\alpha=10^{-4}$ while the parameters of the generator was fixed.
Finally in the third phase, the actual GAN was trained with 8 updates of the discriminator per one update of the generator 
and with $\lambda_1=1$, $\lambda_2=1000$, $\lambda_3=400$,  learning rates $2 \times 10^{-5}$ and $1 \times10^{-4}$ for the generator and discriminator, respectively.  
For a baseline model, the ContextVP, more specifically ContextVP4-WD-big, was trained for the next-frame prediction given an input of 10 past frames 
on two NVIDIA Tesla V100 GPUs with a batch size of 8 and the same optimization parameters with \cite{Byeon_2018_ECCV},  
with the additional flip-flop data augmentation as above mentioned.

\subsection{Metrics}
We adopt Structural Similarity Index Measure (SSIM) \cite{wang2004image} to measure the prediction accuracy. 
However, it is well-known that this metric correlates poorly with human perceptual judgement of visual similarity, and tends to prefer blurriness to naturalness. 
Therefore, we also include the Learned Perceptual Image Patch Similarity (LPIPS) \cite{zhang2018unreasonable}, 
a metric based on the linearly weighted cosine distance of visual features of the pre-trained CNNs, 
which has been shown to correlate better with human perception than SSIM.
More specifically, LPIPS is calculated by using the AlexNet \cite{krizhevsky2012imagenet} as the pre-trained CNN architecture with the $\mathbf{lin}$ configuration.

\setlength{\tabcolsep}{4pt}
\begin{table} 
\begin{center}
\caption{Evaluation of next frame prediction on the Caltech Pedestrian dataset. 
All models are trained for next-frame prediction on the KITTI dataset. 
As an input for the prediction, all the model except for ours take 10 frames, and ours use 8 frames.
Higher values of SSIM, lower values of LPIPS indicate better results. 
({\dag}) This score is from \cite{Liang_2017_ICCV}. 
(+) This score is computed by us using their trained network.  
(*) This score is calculated by us using our implementation of their best model. 
}
\label{table:headings}
\begin{tabular}{lcc}
\hline\noalign{\smallskip}
Method & SSIM & LPIPS ($\times$100)\\
\noalign{\smallskip}
\hline
\noalign{\smallskip}
%Copy-Last-Frame  &  0.779 & 5.11 \\
Copy-Last-Frame  &  0.775 & 5.23 \\
\hline
BeyondMSE \cite{Mathieu16}{\dag}& 0.881 & --- \\
PredNet\cite{Lotter17}+ & 0.906 & 7.83\\
DualMotionGAN\cite{Liang_2017_ICCV} & 0.899 & --- \\
ContextVP\cite{Byeon_2018_ECCV}* & {\bf 0.924}  & 5.14\\
DPG\cite{Gao_2019_ICCV} & 0.923 & 5.04 \\
\hline
Ours(GAN-VGG) & 0.916 & 3.61\\
Ours(G-VGG) & 0.917 & {\bf 3.52}\\
Ours(GAN-MAE) & 0.923 & 4.09\\
Ours(G-MAE) & 0.923 & 4.30\\
\hline
\end{tabular}
\label{tab:next}
\end{center}
\end{table}
\setlength{\tabcolsep}{1.4pt}

\begin{figure}
\centering
\includegraphics[height=14cm]{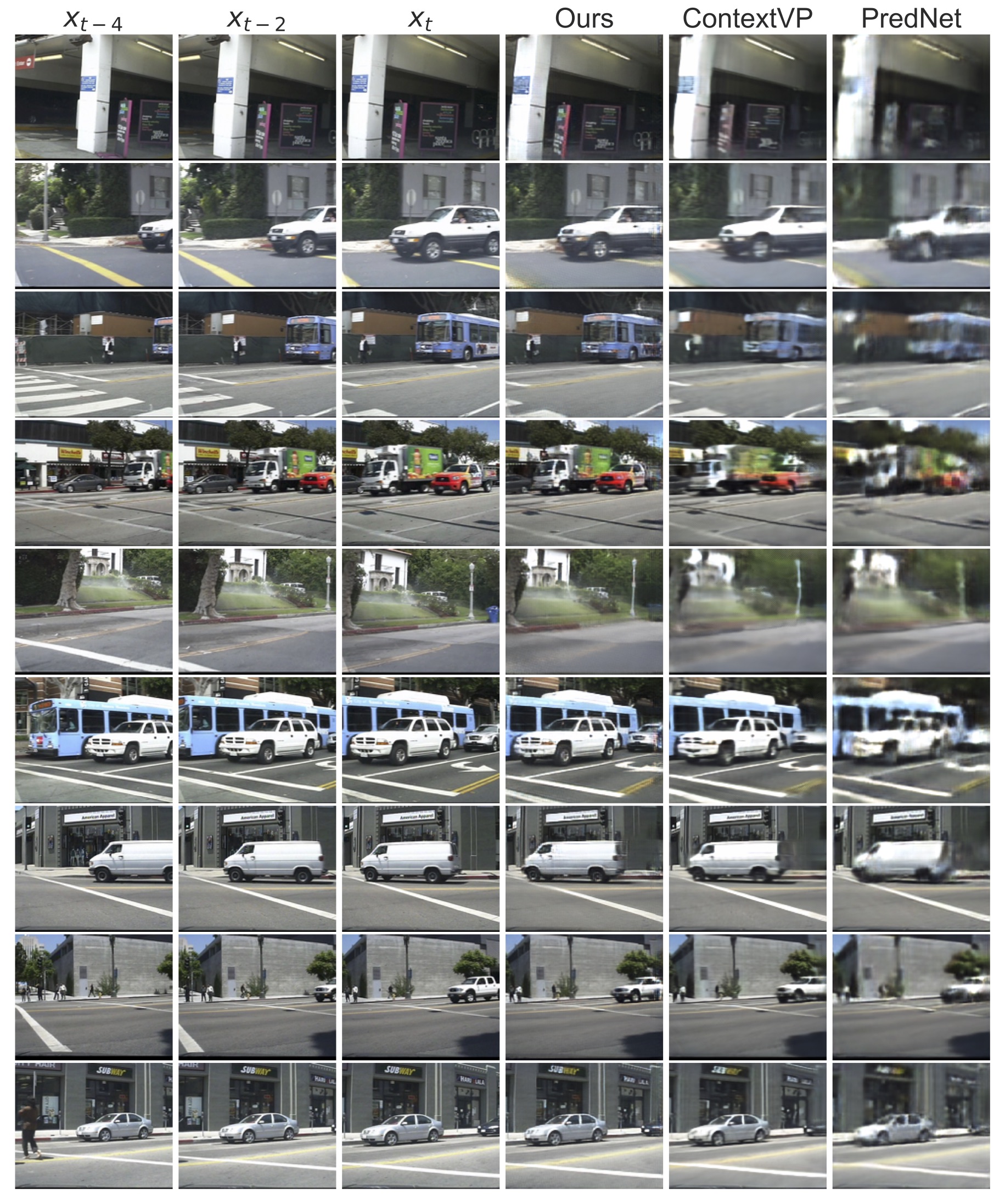}
\caption{Qualitative comparisons of predicted next frames from the Caltech Pedestrian dataset 
among our model (GAN-VGG), and the state-of-the-art models (ContextVP, PredNet). 
All models are trained for next-frame prediction on the KITTI dataset, and tested on the Caltech Pedestrian dataset.
As an input for the prediction, ContextVP and PredNet take 10 frames, and ours use 8 frames. 
Note that $x_t$ indicates the Ground Truth.
Our model generates much sharper results with finer details and textures, and less distorted appearance of object compared to other models.
Results are best viewed in color with zoom
}
\label{fig:next-frame-examples}
\end{figure}

\subsection{Next-Frame Prediction}
We compare our approach with the baseline models, ContextVP \cite{Byeon_2018_ECCV} and PredNet \cite{Lotter17}, 
as well as the BeyondMSE \cite{Mathieu16}, DualMotionGAN \cite{Liang_2017_ICCV}, and DPG \cite{Gao_2019_ICCV}.
We also include the Copy-Last-Frame that simply copies the last input frame. 
As shown in Table \ref{tab:next}, our GAN-based model (GAN-VGG) outperforms the state-of-the-art on LPIPS score, 
while achieving slightly lower SSIM score than ContextVP \cite{Byeon_2018_ECCV} and DPG \cite{Gao_2019_ICCV}.
Samples of results of next-frame prediction from the test set are shown in Fig. \ref{fig:multi-frame-examples} and Fig. \ref{fig:next-frame-examples}. 
Our model suffers less from blurriness and distortions, and produces sharper predictions with finer details than our baselines, 
confirming the previous findings that LPIPS correlates better with human perception than SSIM \cite{zhang2018unreasonable}. 
% artifacts observed at inferred appearance of regions coming into a frame after the last input frame
%     in the case of ContextVP and PredNet, ...  blurry ... distortion
%     Ours:  hatch noise?

\begin{figure}%[htbp]
\centering
\includegraphics[height=6.0cm]{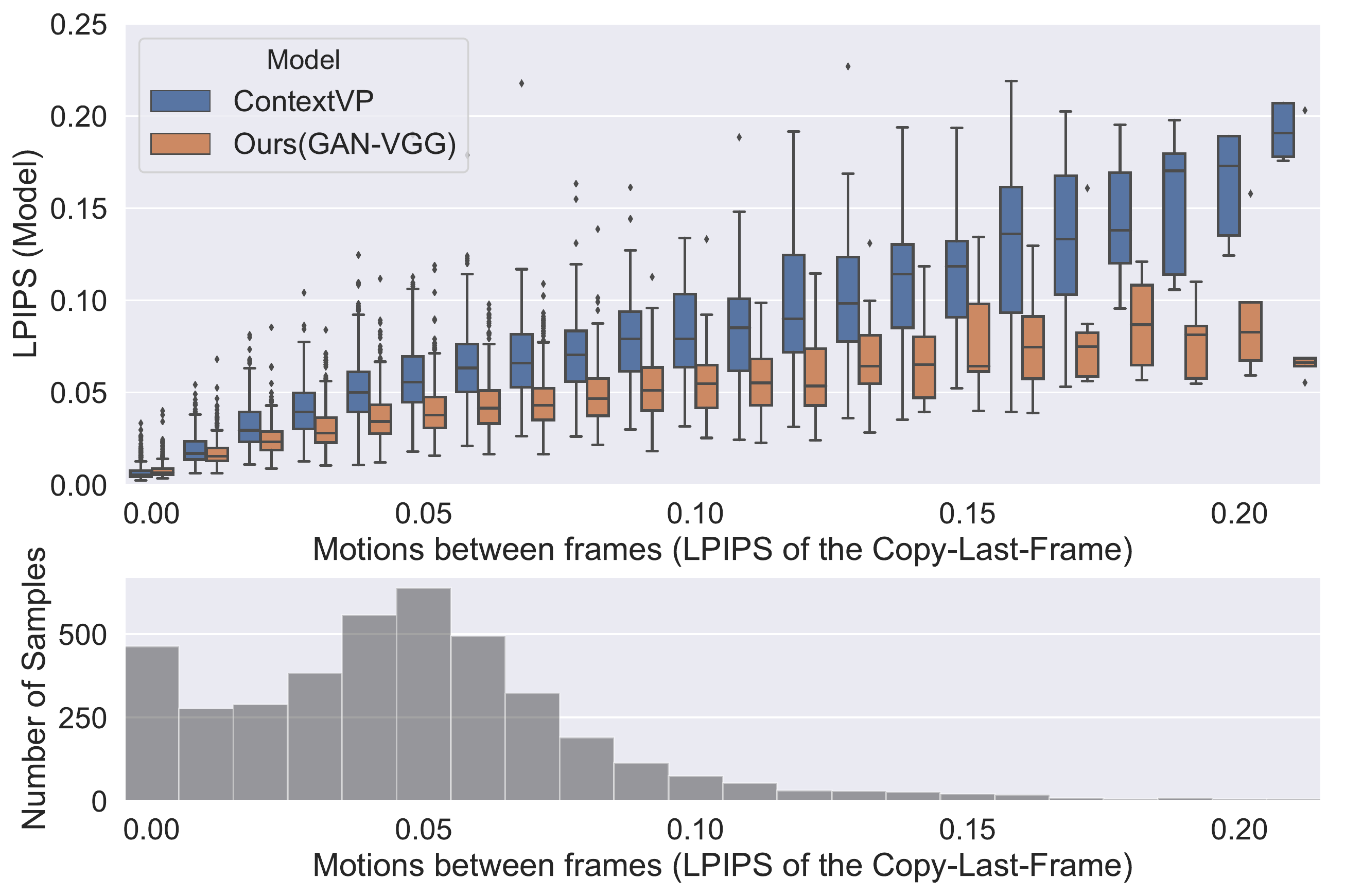}
\caption{
Breakdown of model performance in next-frame prediction on the Caltech Pedestrian dataset based on motions between frames of sample sequences (top), 
and distribution of motions in the Caltech Pedestrian dataset (bottom).
We adopt the LPIPS scores of the Copy-Last-Frame to measure motions in sample sequences. 
Note that the bottom-axis is truncated at 0.22 because we observed at most a single sample per bin above this value.
Results are best viewed in color
}
\label{fig:boxplot}
\end{figure}

We further investigate the relationships between the accuracy of next-frame prediction of models and the degree of motions between frames of sample sequences. 
We use the LPIPS scores of the Copy-Last-Frame to measure the motions of sample sequences. 
Fig. \ref{fig:boxplot} shows that our model is better than the baseline model over a broad range of motions. 
Although major components of the Caltech Pedestrian dataset show relatively slight changes between frames as discussed in \cite{Gao_2019_ICCV} (Table \ref{tab:next}, Fig. \ref{fig:boxplot} \it bottom\rm), 
the dataset also contains challenging sample sequences with large motions between frames.
You can find such challenging samples in Fig. \ref{fig:multi-frame-examples} and Fig. \ref{fig:next-frame-examples},
whose motions range between 0.16 and 0.4.
%In the range of motions between frames, our model outperforms the baseline by increasing margins.
It is the range of motions between frames in which our model outperforms the baseline by increasing margins.
\begin{figure}%[htb]
\centering
\includegraphics[height=6.5cm]{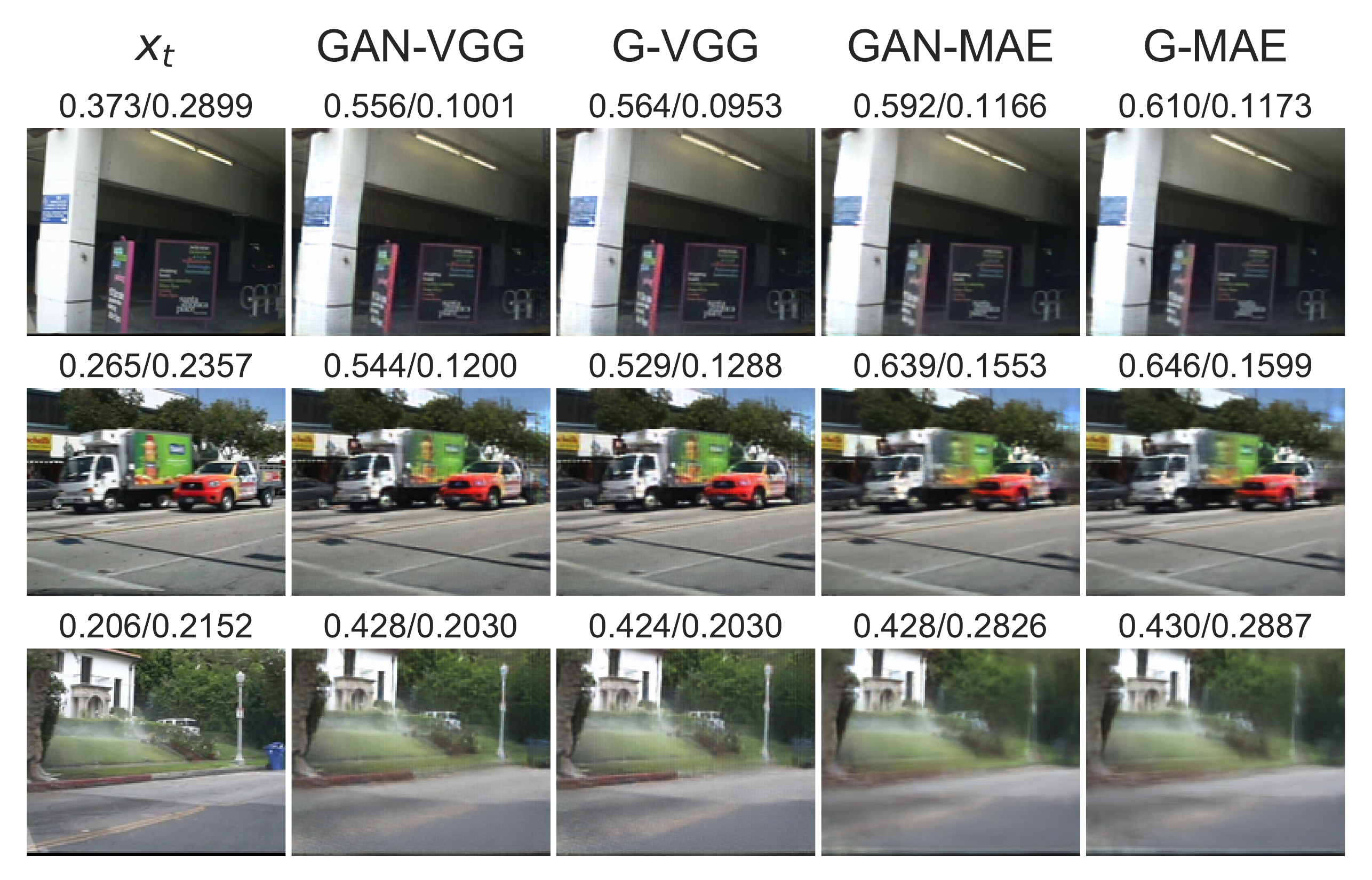}
\caption{
Qualitative comparisons of predicted next frames of our model variants with different choices of loss functions. 
We report SSIM/LPIPS for the example image. Note that SSIM/LPIPS of $x_t$ (ground truth) is of the Copy-Last-Frame. Results are best viewed in color with zoom
}
\label{fig:compare_models_samples}
\end{figure}

Moreover, we investigate the effect of different choices of loss functions for the GAN-based networks. 
Specifically, we investigate the GAN-based networks with the standard MAE loss but not with VGG loss (GAN-MAE).
We also evaluate the performance of the generator network optimized for the two losses without the adversarial loss function $\mathcal{L}_{adv}^G$, 
$\mathcal{L}_{MAE}^G = \lambda_2 \mathcal{L}_{MAE}$  (G-MAE) and 
$\mathcal{L}_{VGG}^G = \lambda_2 \mathcal{L}_{MAE}+ \lambda_3 \mathcal{L}_{VGG}$ (G-VGG).
Quantitative results are summarized in Table \ref{tab:next}, % and Fig. \ref{fig:boxplot_btw_models}, 
and visual examples are provided in Fig. \ref{fig:compare_models_samples}.
The MAE loss by itself provides solutions with the highest SSIM  scores which are comparable to, 
and with better LPIPS scores than those of the state-of-the-art models, indicating the advantage of our design of network architecture over baseline models.
However, results of the solutions are perceptually rather blurry and smooth than results with the VGG loss, 
even combined with the adversarial loss which further improves the LPIPS score. 
Our model variants which include both VGG and MAE loss functions (GAN-VGG, G-VGG) outperform other model variants without VGG loss (GAN-MAE, G-MAE) and the baseline models on the LPIPS score.  
G-VGG achieves the new state-of-the-art on the Caltech Pedestrian dataset in terms of LPIPS, slightly outperforming GAN-VGG. 
Visual results of G-VGG are sharper and have finer details and textures than results of the model variants without VGG loss, however, are perceptually indistinguishable from results of GAN-VGG.
These results indicate that the VGG loss significantly contributes for improving next-frame predictions while the adversarial loss is helpful when without the VGG loss.

\begin{figure}%[htb]
\centering
\includegraphics[height=4.5cm]{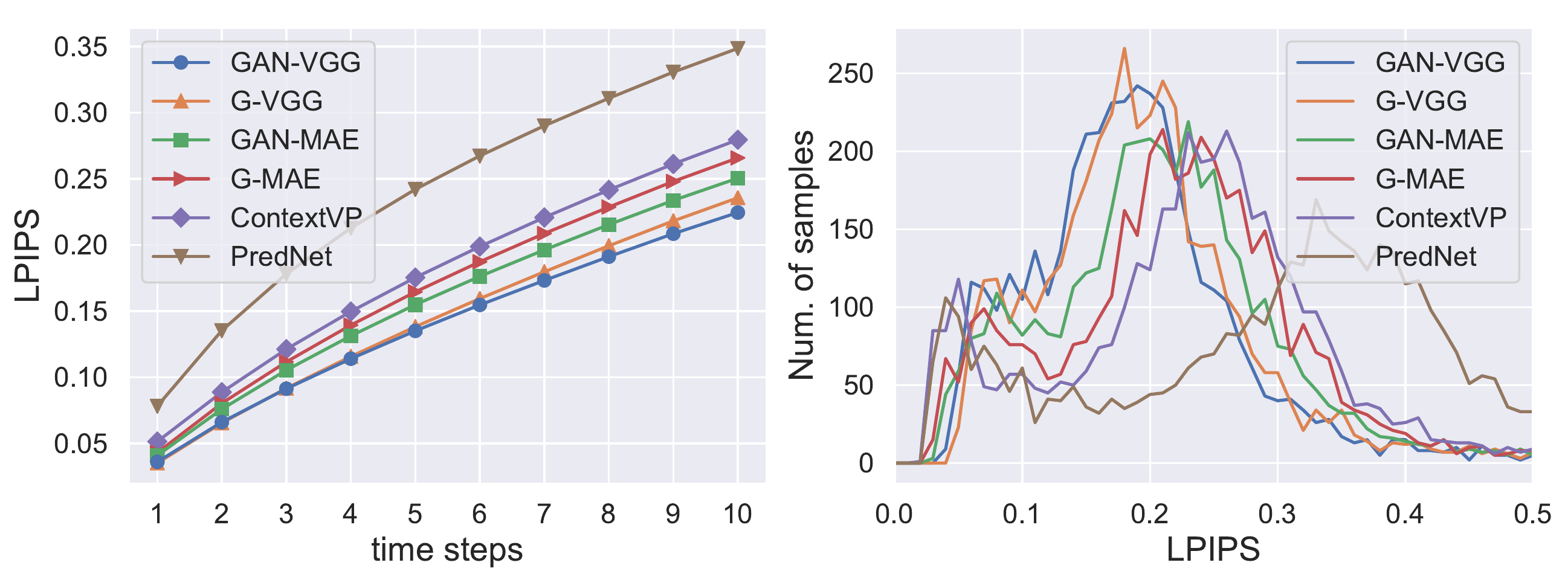}
\caption{
Comparison of multi-step prediction on the Caltech Pedestrian dataset (left) and distributions of LPIPS scores at the 9-th future step (right):
 our models (GAN-VGG, G-VGG, GAN-MAE, G-MAE), and baseline models (ContextVP, PredNet).
Baseline models and our models are trained for next-frame prediction on the KITTI dataset, 
and tested to recursively predict 10 future frames on the Caltech Pedestrian dataset.
%As an input for the prediction, baseline models and our models take 10 and 8 frames, respectively. 
Results are best viewed in color
}
\label{fig:compare_multi_step_lpips}
\end{figure}
\begin{figure}% [htb]
\centering
\includegraphics[height=16.0cm]{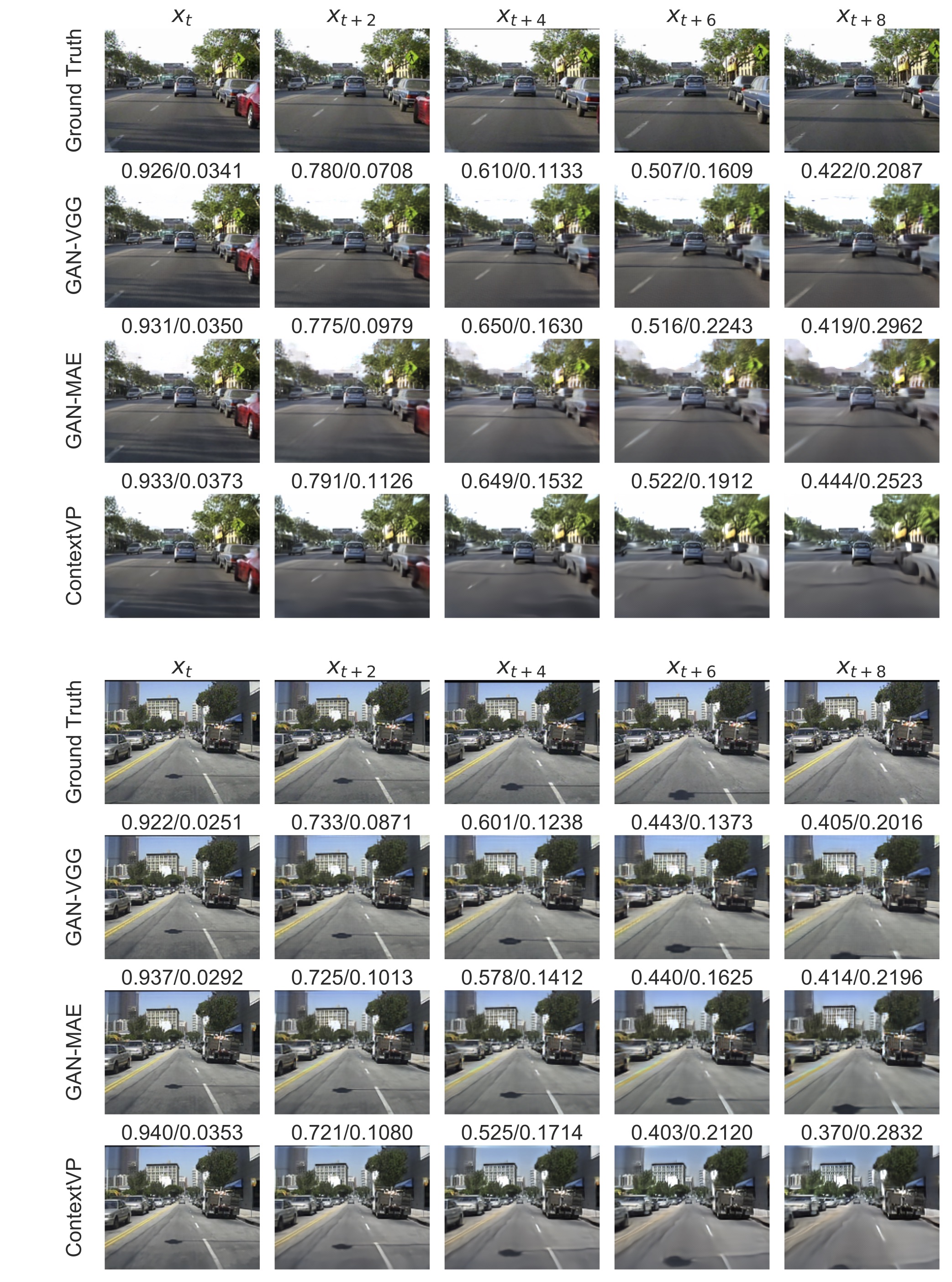}
\caption{
Qualitative comparison of predicted frames between the state-of-the-art model, ContextVP \cite{Byeon_2018_ECCV}, and our models (GAN-VGG, GAN-MAE). 
ContextVP and our models are trained for next-frame prediction on the KITTI dataset, 
and tested to recursively predict 10 future frames on the Caltech Pedestrian dataset.
As an input for the prediction, ContextVP and our models take 10 and 8 frames, respectively. 
We report SSIM/LPIPS for the example image. 
%Note that SSIM/LPIPS of $x_t$ (ground truth) is of the Copy-Last-Frame. 
Results are best viewed in color with zoom
}
\label{fig:compare_multi_step_samples}
\end{figure}
\subsection{Multi-Step Prediction}
Fig. \ref{fig:compare_multi_step_lpips} compares multi-step prediction results of our models with the baseline models, ContextVP and PredNet. 
All models are trained for next-frame prediction on the KITTI dataset, 
and tested to recursively predict 10 future frames on the Caltech Pedestrian dataset.
As an input for the prediction, baseline models and our models take 10 and 8 frames, respectively. 
Like the results of next-frame prediction, our models with the combination of the VGG and MAE loss functions (GAN-VGG, G-VGG) are better 
than the baseline models and our model variants without the VGG loss function (GAN-MAE, G-MAE). 
In contrast to the next-frame prediction, GAN-VGG is better than G-VGG when predicting further future frames. 
The distributions of LPIPS scores at the 9-th future step are bimodal (Fig. \ref{fig:compare_multi_step_lpips}, right).
These two peaks correspond to samples of very small motions ($<0.01)$ and larger motions, respectively.
Our models are better than the baselines with samples of larger motions, but not with samples of very small or no motions as in the case of next-frame prediction (leftmost bin in Fig. \ref{fig:boxplot}).
Quantitative comparisons with samples of average and large motions are shown in Fig. \ref{fig:compare_multi_step_samples} and Fig. \ref{fig:multi-frame-examples}, respectively.
Although appearances of objects in the predicted frames are getting blurred and distorted with recursive predictions, GAN-VGG suffers less from these artifacts. 
%As obvious in recursive predictions on samples of large motions, ... highly uncertain regions caused by fast camera movements ... inferring ... based on neighboring patches ...
% large motions: 
%   multi-frame-predictions_02468_32_1168.pdf, multi-frame-predictions_02468_541_1849.pdf, multi-frame-predictions_02468_1071_3894.pdf
% average motions:
%   multi-frame-predictions_02468_595_2211.pdf, multi-frame-predictions_02468_812_2211.pdf, 
%   multi-frame-predictions_02468_2211_3782.pdf, multi-frame-predictions_02468_3782_3901.pdf

\section{Conclusions}
We have described deep hierarchical residual network models for photo-realistic video prediction (G-VGG and GAN-VGG) that set new state of the arts of next-frame and multi-step predictions, respectively,  
on the public dataset of car-mounted camera videos when evaluated with LPIPS measure.
They generate sharp future frames with finer details and textures that are more perceptually realistic than the baseline models, especially under fast and large camera movements.
Ablation studies with various settings of loss functions indicate that the perceptual loss function is very effective for generating photo-realistic predictions 
and that the adversarial loss function further improves multi-step predictions.

%This is the last page of the manuscript.
%\par\vfill\par
%Now we have reached the maximum size of the ECCV 2020 submission (excluding references).
%References should start immediately after the main text, but can continue on p.15 if needed.

%\clearpage
% ---- Bibliography ----
%
% BibTeX users should specify bibliography style 'splncs04'.
% References will then be sorted and formatted in the correct style.
%
\bibliographystyle{splncs04}
%\bibliography{egbib}
\bibliography{reference}
\end{document}